\documentclass[letterpaper]{article} 
\usepackage[submission]{aaai23}  
\usepackage{times}  
\usepackage{helvet}  
\usepackage{courier}  
\usepackage[hyphens]{url}  
\usepackage{graphicx} 
\usepackage{natbib}  
\usepackage{caption} 
\usepackage{algorithm}
\usepackage{algorithmic}
\usepackage{xcolor}
\usepackage{amsmath}
\usepackage{makecell}
\usepackage{amsmath,amsfonts}
\usepackage{algorithmic}
\usepackage{algorithm}
\usepackage{array}
\usepackage[caption=false,font=normalsize,labelfont=sf,textfont=sf]{subfig}
\usepackage{textcomp}
\usepackage{stfloats}
\usepackage{url}
\usepackage{verbatim}
\usepackage{float}
\usepackage{cite}
\usepackage{tabu}                     
\usepackage{multirow}                 
\usepackage{multicol}                 
\usepackage{multirow}                
\usepackage{float}                    
\usepackage{makecell}                 
\usepackage{booktabs}                 

\usepackage{newfloat}
\usepackage{listings}
\DeclareCaptionStyle{ruled}{labelfont=normalfont,labelsep=colon,strut=off} 
\floatstyle{ruled}
\newfloat{listing}{tb}{lst}{}
\floatname{listing}{Listing}
\title{MDFL: Multi-domain Diffusion-driven Feature Learning}
\author{
    Written by AAAI Press Staff\textsuperscript{\rm 1}\thanks{With help from the AAAI Publications Committee.}\\
    AAAI Style Contributions by Pater Patel Schneider,
    Sunil Issar,\\
    J. Scott Penberthy,
    George Ferguson,
    Hans Guesgen,
    Francisco Cruz\equalcontrib,
    Marc Pujol-Gonzalez\equalcontrib
}
\affiliations{
    \textsuperscript{\rm 1}Association for the Advancement of Artificial Intelligence\\

    1900 Embarcadero Road, Suite 101\\
    Palo Alto, California 94303-3310 USA\\
    publications23@aaai.org

}

\usepackage{bibentry}

\begin{document}

\maketitle

\begin{abstract}
High-dimensional images, known for their rich semantic information, are widely applied in remote sensing and other fields. The spatial information in these images reflects the object's texture features, while the spectral information reveals the potential spectral representations across different bands. Currently, the understanding of high-dimensional images remains limited to a single-domain perspective with performance degradation. Motivated by the masking texture effect observed in the human visual system, we present a multi-domain diffusion-driven feature learning network (MDFL) , a scheme to redefine the effective information domain that the model really focuses on. This method employs diffusion-based posterior sampling to explicitly consider joint information interactions between the high-dimensional manifold structures in the spectral, spatial, and frequency domains, thereby eliminating the influence of masking texture effects in visual models. Additionally, we introduce a feature reuse mechanism to gather deep and raw features of high-dimensional data. We demonstrate that MDFL significantly improves the feature extraction performance of high-dimensional data, thereby providing a powerful aid for revealing the intrinsic patterns and structures of such data. The experimental results on three multi-modal remote sensing datasets show that MDFL reaches an average overall accuracy of 98.25$\%$, outperforming various state-of-the-art baseline schemes. The code will be released, contributing to the computer vision community.
\end{abstract}

\section{Introduction}
\label{}

Recently, remarkable strides have been achieved in precise classification within the realm of natural images, partially attributing to the distinctive representations found within the ImageNet dataset \cite{li2022bigdatasetgan}.  These representations frequently encompass local structures, repetitive patterns, and hierarchical arrangements \cite{yao2019hierarchy}, rendering them advantages for deep learning models. However, progress in the domain of high-dimensional data analysis has been relatively sluggish in comparison to that in natural images due to several factors, including the absence of prior knowledge, sparse distributions, and low spatial resolutions \cite{jiang2021lren}. Consequently, it becomes imperative to delve into exploring intelligent interpretation algorithms that can be effectively applied to high-dimensional data, thereby improving their classification accuracy and effectiveness.

\begin{figure}[t]
	\centering
	\includegraphics[width=3.4in]{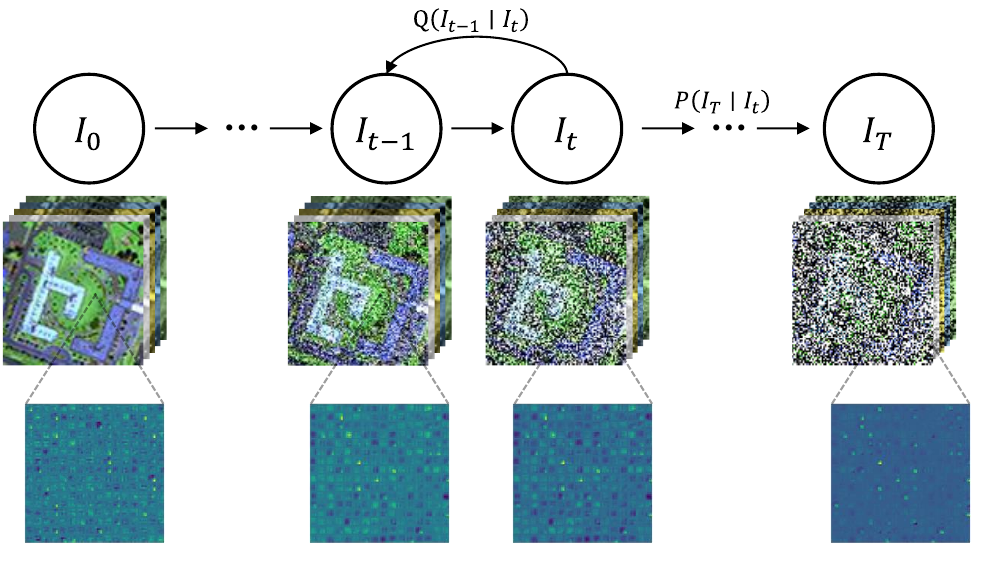}
	\caption{The forward diffusion process in the spectral-spatial domain denoted by $I_0$, the noised image at time step $0$ and time step $T$ denoted by $I_T$, and the feature visualization results of the hyperspectral image at different time steps.}
	\label{fig1}
	\vspace{-0.2in}
\end{figure}

Multi-modal learning (MML), as a vital endeavor in high-dimensional data processing, plays a pivotal role in remote sensing applications \cite{zhao2023ddfm} by harnessing essential information from multiple sensor source images to enhance data acquisition \cite{feng2023cross,hong2020more}. In scenarios where the availability of prior samples is limited, researchers have begun exploring fusion analysis approaches for high-dimensional data. Many of these approaches leverage Convolutional Neural Network (CNN) architectures for inter-layer fusion. For instance, Hong \emph{et al.} \cite{9179756} proposed a deep encoder-decoder network architecture for hyperspectral and LiDAR data classification, effectively capturing spectral and spatial information in a deep latent space. Subsequently, larger-scale models like transformers \cite{roy2023multimodal} and diffusion model \cite{zhao2023ddfm} have emerged. Typically, the optimization of these methods involves two distinct steps: extracting features from high-dimensional data using large-scale model architectures, and subsequently classifying the extracted features using a classifier. While these methods mitigate the curse of dimensionality, their feature extraction process does not adequately cater to the manifold structures of high-dimensional data, and result in suboptimal solutions. Moreover, compared to natural images with feature domains in spatial, spectral, and frequency domains, high-dimensional data often exhibit more compact data relationships \cite{hu2022hdnet}. Therefore, the majority of current techniques that exhibit exceptional performance in natural image scenarios are not applicable in high-dimensional contexts.

The aforementioned methodologies approach the fusion of multi-modal data as a mere feature extraction problem. Such methods fail to fully exploit the particularity of multi-modal fusion, which involves preserving the spectral, spatial, and frequency domains inherent in high-dimensional data. Thus, they treat the fusion process as a black-box deep learning problem. Motivated by the masking texture effect observed in the human visual system \cite{liu2021human}, the bottom-up attention driven by objective content can locate and analyze important target domains in detail, while other regions are only roughly analyzed or even ignored, this characteristic effectively enhances the focusing ability of image content and improves the performance accuracy of target recognition. However, excessive attention to a single target domain will lead to the loss of analytical capabilities for other effective information domains in the image \cite{liang2022dine}, which is consistent with the texture masking effect \cite{song2020bi,li2022mat}. This observation will be validated in the subsequent multi-domain learning methods. As shown in Figure \ref{fig1}, a spectral-spatial sample of multi-modal data is obtained in the preceding process of diffusion. However, with the increase of time steps, the texture occlusion effect caused by the digital domain information makes it difficult to extract effective features. For the multi-modal fusion problem, it is obvious that preserving the spatial, spectral, and frequency domains is the main objective. Therefore, deep learning methods should explicitly focus on this aspect, which inspires our proposed deep network called “MDFL”. To the best of our knowledge, this is the first discussion on diffusion characteristics in the frequency domain of high-dimensional data, and the first exploration of the intrinsic properties of high-dimensional data by combining spatial and spectral information:

\begin{itemize}
	\item We propose a diffusion-driven posterior sampling model MDFL, which formulates the MML problem as a unified framework for multi-domain joint feature learning in a manifold structure. It significantly improves feature extraction performance for high-dimensional data.
	\item A novel feature reuse mechanism is developed to integrate learned deep and shallow features. Consisting of two parallel attention modules, it can effectively aggregate cross-level features with minimal additional cost.
	\item We demonstrate the superiority of the proposed approach over the existing SOTA in multiple multimodal datasets. In addition, the effectiveness of feature reuse mechanism and multi-domain learning for high-dimensional data interpretation is verified by ablation experiments.
\end{itemize}

\section{Related Works}
\label{Related}
\subsection{Diffusion Models}
The diffusion model is based on non-equilibrium thermodynamics and involves a Markov diffusion step chain that adds random noise to the data and learns the reverse diffusion process to generate desired data samples from the noise \cite{croitoru2023diffusion}. Unlike VAE or flow models, diffusion models have a fixed process and high-dimensional latent variables \cite{nichol2021improved}. Diffusion models have been applied to various fields such as image super-resolution \cite{esser2021taming}, semantic segmentation \cite{baranchuk2021label}, and classification \cite{han2022card}. Research on these can be categorized into three areas: effective sampling, improved likelihood estimation, and handling data with special structures \cite{yang2022diffusion}. Effective sampling involves generating samples using iterative methods with a large number of evaluation steps. Karras \emph{et al.} \cite{karras2022elucidating} determined optimal time discretization and applied high-order Runge-Kutta methods for sampling. They also evaluated different sampler schedules and analyzed the role of randomness in the sampling process. Diffusion models are trained using the variational lower bound (VLB) of the log-likelihood. Remote sensing data, which include observations from satellites or sensors, often exhibit spatial correlations and geographic distributions. Zhou \emph{et al.} \cite{zhou2023hyperspectral} designed a temporal leap feature library and dynamic feature fusion module to utilize rich temporal leap features and learn information-rich multi-temporal representations. Han \emph{et al.} \cite{han2022card} introduced the classification and regression diffusion model, which combines a denoising diffusion-driven generative model with a pretrained mean estimator for more accurate instance-level confidence evaluations in classification tasks.

High-dimensional hyperspectral images have been widely used as a representative type of data. In order to fully utilize the digital domain characteristics of hyperspectral data, diffusion-based classification models have been proposed. Although these methods have achieved satisfactory results, challenges still persist in the fusion of multi-modal data, such as the inability to capture the close relationships among multiple target domains. Additionally, current diffusion models used for remote sensing classification do not consider the manifold structure of high-dimensional data. To better exploit the information in the spectral domain and spectral-spatial domain, we propose a diffusion-driven multi-domain learning method and verify its effectiveness through extensive comparative experiments.

\section{Methodology}

\subsection{Problem Formulation}
The problem of high dimensional land cover classification can be defined as accurately assigning each pixel in a remote sensing image to different land cover classes. Given a high dimensional image $I$ composed of $p$-th pixels, we focus on how to train different modalities of data $I^1, I^2 \in \mathbb{R}^{h \times w \times c}$ to achieve classification tasks. The $h$, $w$, and $c$ denote the height, width, and the number of image channels, and $I^1(p)$ is represented as the $p$-th pixel pair of the first modality. The two modalities capture the same scene with the same label information, denoted as ${L}\in\mathbb{R}^{h \times w \times c}$ with $C\in N^*$ classification categories, to represent various land cover categories, such as buildings, roads, and fields. The objective of image classification is to develop a model  $\Theta_{\left(I^1, I^2\right)}$, that can effectively map input images from various modalities to a novel representation $C_{\max }\left(I^1, I^2\right)$, thus, indicating the probability of each pixel being associated with different categories. By setting the maximum probability of distinct categories as the threshold ($\tau$), a binary prediction map is obtained through hard classification, The values $1$ and $0$ in the map respectively represent the presence of the specific category and other categories, which is defined as: 

\begin{equation}
	\Theta_{\left(I^1, I^2\right)} = \begin{cases}0, & \text { if } C_{\max }\left(I^1, I^2\right)<\tau, \\ 1, & \text { otherwise. }\end{cases}
\end{equation}

\begin{figure*}[t]
	\centering
	\includegraphics[width=5.5in]{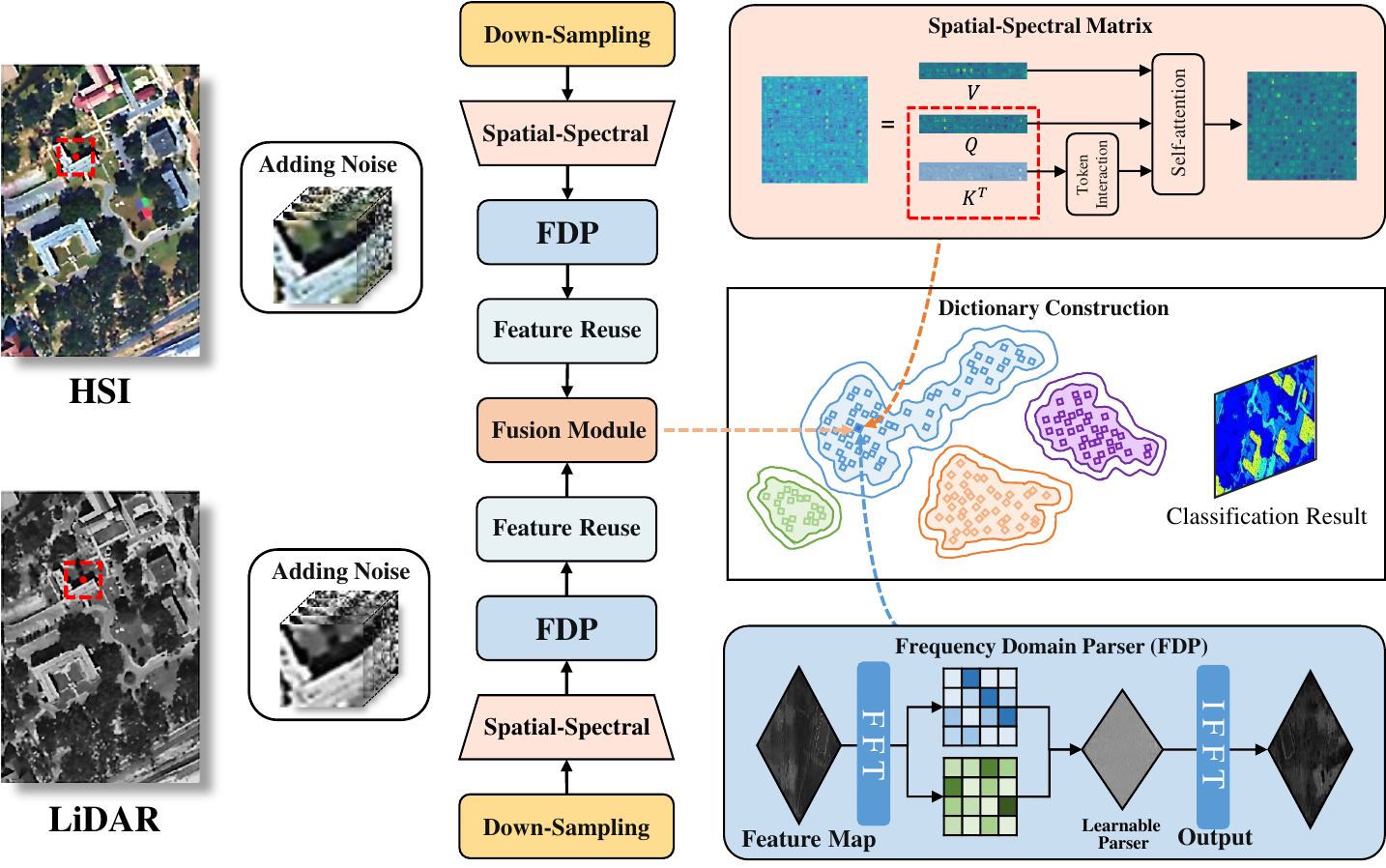}
	\caption{Overview of the multi-domain diffusion-driven feature learning framework, which combines spatial-spectral matrix and frequency domain parser for joint learning of multi-modal data. The MDFL framework is utilized to integrate the two modal data at the feature level, after which the extracted features are categorized into the downstream model. }
	\label{framework}
\end{figure*}

Based on this basic model, we propose the MDFL network architecture. As shown in Figure \ref{framework}, the network consists of two branches that are multi-domain learning driven by a diffusion model. Thus, the fusion future graph in our work $C_{\max}\left(I^1, I^2\right)$ can be expressed as:

\begin{equation}
	C_{\max }\left(I^1, I^2\right)=\Phi\left(I^1, I^2 \mid \alpha_1, \alpha_2\right),
\end{equation}
where a nonlinear objective model $\Phi(\cdot)$ is used to transform the image space into the classification space, and $\alpha_1, \alpha_2$ represent the corresponding parameters of the two branches.

\subsection{Network Architecture}
In this section, the network architecture of the diffusion-driven multi-domain feature fusion learning method is introduced in Figure \ref{framework} to explicitly model the compact representation of each modal target domain for fusion classification of multi-modal data.

\subsubsection{Diffusion-driven spectral-spatial feature learning}
In the realm of high-dimensional data analysis, a thorough investigation of the spectral and spatial domains is crucial for elucidating the diffusion characteristics. This section proposes a novel approach to learn diffusion-driven spectral-spatial features in a multi-domain posterior sampling framework. The proposed method utilizes a conditional generation module and a maximum likelihood estimation module, the two jointly analyzing the spectral, spatial and frequency domains. First, we add noise to the digital domain instance in the forward propagation stage and take the fusion of multiple noise scales as the input of the model. As shown in Figure \ref{fig1}, the diffusion process of the preceding term is inspired by the principle of non-equilibrium thermodynamics. As a Markov chain, the one-step derivation formula can be obtained from the iterative formula:

\begin{equation}
	\left\{\begin{array}{l}
		{I}_{\boldsymbol{t}}=\sqrt{\bar{\alpha}_t} {I}_{\mathbf{0}}+\sqrt{1-\bar{\alpha}_t} \boldsymbol{\epsilon}, \\
		\bar{\alpha}_t=\prod_{i=1}^t \alpha_i,
	\end{array}\right.
\end{equation}
where $\overline{\alpha_t}=\prod_i^t \alpha_i$ is the hyperparameter set with the Noise schedule, and $\epsilon \sim N(0,1)$ is Gaussian noise.
The noised maps are firstly concatenation in channel dimension as:
\begin{equation}
	\widetilde{I}=\operatorname{Concat}\left[{I}_{0},{I}_{50},{I}_{100},{I}_{200},{I}_{400}\right].
\end{equation}

During training, a U-Net basic framework $f_\theta\left(\boldsymbol{z}_t, t\right)$ is trained to predict $\widetilde{I}$ from $\boldsymbol{z}_t$ by minimizing the training objective with $\ell_2$ loss:
$$
\mathcal{L}_{\text {train }}=\frac{1}{2}\left\|~f_\theta\left(\boldsymbol{z}_t, t\right)-\widetilde{I}~\right\|_2,
$$
based on Bayes' theorem, it is found that the posterior $q\left(\boldsymbol{z}_{t-1} \mid \boldsymbol{z}_t, \boldsymbol{z}_0\right)$ is a Gaussian distribution as well:
\begin{equation}
	q\left(\boldsymbol{z}_{t-1} \mid \boldsymbol{z}_t, \boldsymbol{z}_0\right)=\mathcal{N}\left(\boldsymbol{z}_{t-1} ; \tilde{\mu}\left(\boldsymbol{z}_t, \boldsymbol{z}_0\right), \tilde{\beta}_t \widetilde{I}\right),
\end{equation}
where
\begin{equation}
	\tilde{\mu}_t\left(\boldsymbol{z}_t, \boldsymbol{z}_0\right) = \frac{\sqrt{\bar{\alpha}_{t-1}} (1-{\alpha}_t)}{1-\bar{\alpha}_t} \boldsymbol{z}_0+\frac{\sqrt{\alpha_t}\left(1-\bar{\alpha}_{t-1}\right)}{1-\bar{\alpha}_t} \boldsymbol{z}_t,
\end{equation}
and
\begin{equation}
	\tilde{\beta}_t = \frac{1-\bar{\alpha}_{t-1}}{1-\bar{\alpha}_t} (1-{\alpha}_t)
\end{equation}
are mean and variance of this Gaussian distribution.
We could get a sample from $q\left(\boldsymbol{z}_0\right)$ by first sampling from $q\left(\boldsymbol{z}_T\right)$ and running the reversing steps $q\left(\boldsymbol{z}_{t-1} \mid \boldsymbol{z}_t\right)$ until $\boldsymbol{z}_0$. Besides, the distribution of $q\left(\boldsymbol{z}_T\right)$ is nearly an isotropic Gaussian distribution with a sufficiently large $T$ and reasonable schedule of $\beta_t\left(\beta_t \rightarrow 0\right)$, which makes it trivial to sample $\boldsymbol{z}_T \sim \mathcal{N}(0, \widetilde{I})$. Moreover, we could approximate $q\left(\boldsymbol{z}_{t-1} \mid \boldsymbol{z}_t\right)$ using a neural network, due to the fact that calculating $q\left(\boldsymbol{z}_{t-1} \mid \boldsymbol{z}_t\right)$ exactly should depend on the entire data distribution. The network is optimized to predict a mean $\mu_\theta$ and a diagonal covariance matrix $\Sigma_\theta$ :
\begin{equation}
	p_\theta\left(\boldsymbol{z}_{t-1} \mid \boldsymbol{z}_t\right):=\mathcal{N}\left(\boldsymbol{z}_{t-1} ; \mu_\theta\left(\boldsymbol{z}_t, t\right), \Sigma_\theta\left(\boldsymbol{z}_t, t\right)\right).
\end{equation}
At inference stage, data sample $z_0$ is reconstructed from noise $z_T$ with the model $f_\theta$ both and an updating rule in an iterative way, i.e., $\boldsymbol{z}_T \rightarrow \boldsymbol{z}_{T-\Delta} \rightarrow \ldots \rightarrow \boldsymbol{z}_0$.
At the same time, in order to capture the spatial and spectral residual information in the digital domain of the feature map, we add the spatial-spectral attention mechanism to the UNet basic framework. We then combine the attention mechanism and residual connection to enhance the representation ability of the feature map. 
Furthermore, we construct the transformation matrix ($W_Q$, $W_K$, $W_V$) before the multi-head self-attention (MSA) operation \cite{fan2021multiscale}. 
\begin{equation}
	Q=\widetilde{F}  W_Q, \quad K=\widetilde{F} W_K, \quad V=\widetilde{F} W_V.
\end{equation}

The input intermediate feature $\widetilde{F}$ is partitioned into distinct $Q$, $K$, and $V$ tensors.  To calculate the attention weights, we initially perform a dot product operation between the query and key tensors.  Subsequently, we scale this dot product by a factor of ${1}/{\sqrt{c}}$ to address the issue of vanishing gradients.  Finally, the obtained dot product is passed through a softmax activation function, yielding normalized weights that reflect the relative significance of various elements within the input tensor.
\begin{equation}
	A=\operatorname{Attention}(Q, K, V)=\operatorname{softmax}\left(\frac{Q K^T}{\sqrt{c}}\right)V.
\end{equation}
A linear layer is used to produce the output. Multi-head self-attention splits the queries, keys and values to $h$ parts and performs the attention function in parallel, and then the output values of each head are concatenated and linearly projected to form the final output.

%
%
The resulting tensor is then reshaped to its original dimensions, providing a refined representation of the input tensor capturing relevant spatial and spectral features. This module enables our model to effectively capture and exploit the intrinsic interdependencies between elements in the input tensor, thereby facilitating enhanced feature extraction and improving the interpretation performance of high-dimensional data in the number domain.


\subsubsection{Frequency-aware discriminative feature learning}
\begin{figure}[htbp]
	\centering
	\includegraphics[width=3.26in]{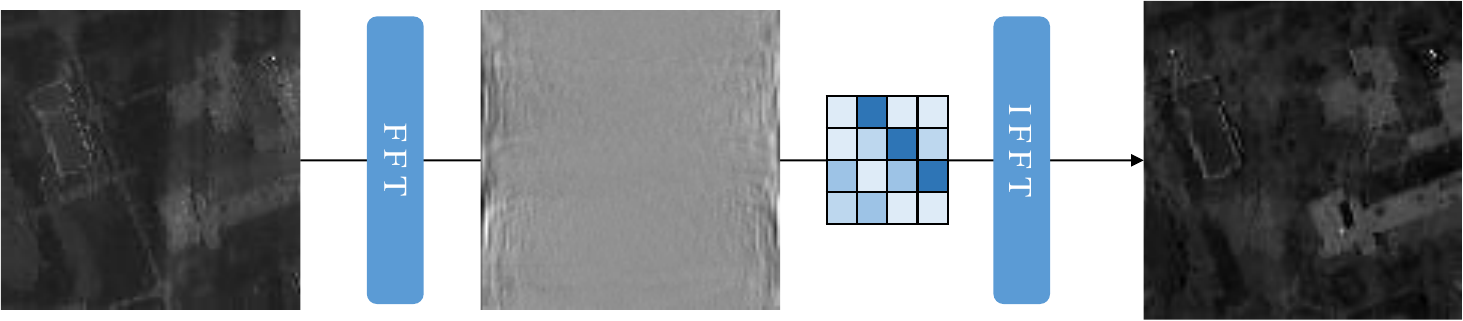}
	\caption{The frequency domain parser visualizations.}
	\label{fft}
\end{figure}

The visual masking effect will be caused with the numerical domain interpretation of only high-dimensional data. As suggested, the phase of a blurry image plays an important role in blurring, providing faithful information about motion patterns \cite{pan2019phase}. The blurry naturally exists in remote sensing images whose resolution is constrained by imaging distance. We introduce the frequency domain to enhance the detail information (such as texture and color information) in the different multi-modalities to make the object more discriminative. The pipeline and effect of frequency-aware discriminative feature learning is shown in Figure \ref{fft}. Our main idea is to learn a parameterized filter by applying it to the Fourier-space features. Let $X(p) \in \mathbb{R}^{H\times W \times C}$ be the input feature matrix, where $H$, $W$, and $C$ indicate the height, width, and channel of the respectively feature. Firstly, the 2D FFT is performed along the spatial dimensions, which can be represented as:
\begin{equation}
	M=\mathcal{F}(X(p)) \in \mathbb{R}^{H \times W \times C},
\end{equation}
where $\mathcal{F}(\cdot)$ denotes the 2D FFT. We then modulate a convolution operation to obtain an attention map that can expose the importance of the different frequency components. In other words, the weight of convolution can be regarded as a learnable version of frequency filters widely utilized in digital image processing.
\begin{equation}
	M^{\prime}=W \otimes M,
\end{equation}
where $W$ denotes the trainable weights in the frequency domain. It serves the purpose of regulating the noise at high frequencies while simultaneously and learning the output of the spectral-spatial convolution from the preceding layer of the network, considering with its spectral characteristics considered. And $\otimes$ represents the element-wise product. We reverse back to the spatial domain by adopting inverse FFT:
\begin{equation}
	X^{\prime}=\mathcal{F}^{-1}(M^{\prime}).
\end{equation}

Different from the spectral-spatial domain, FDFL makes globally adjustments to the components of the specific frequencies to enhance the discriminative information. It can be learned to constrain the different frequency components for adaptive integration.

\subsubsection{Feature reuse module}
In this framework, all network layers share a objective, which is to extract diverse types of features to achieve accurate recognition. The downstream model is expected to leverage the deep features and shallow features from the upstream model U-net to optimize efficiency. To actualize this concept, we introduce a feature reuse mechanism that combines the inherent and semantic information of an image. Specifically, we propose two parallel attention modules, wherein each module modulates the current layer to attain high-level features. This procedure can be described as follows:

\begin{equation}
	\begin{aligned}
		X_{\text {FRM }} & ={ FRM }\left(X_{\text {low}}, X_{\text {deep}}\right) \\
		& =X_{\text {low}}+X_{\text {deep}},
	\end{aligned}
\end{equation}
where $FRM(\cdot)$ represents the mapping function learned by the feature reuse module. $X_{\text{low}}$ and $X_{\text{deep}}$ represent the shallow and deep features in the U-Net encoder, respectively. $X_{\text{low}}$ and $X_{\text{deep}}$ are the attention features in shallow layer and deep layer, which can be obtained as follows:

\begin{equation}
	X_{\text {deep}}=Sigmoid\left(F_d\left(X_{\text {low }}\right)\right) F_b\left(X_{\text {deep}}\right)+F_b\left(X_{\text {low }}\right),
\end{equation}

\begin{equation}
	X_{\text {low}}=Sigmoid\left(F_l\left(X_{\text {low }}\right)\right) F_b\left(X_{\text {deep}}\right)+F_b\left(X_{\text {low }}\right),
\end{equation}
$F_b(\cdot)$ is suitable for the 1$\times$1 convolutional layer structure constraining high-frequency noise, $F_d(\cdot)$ represents the 1$\times$1 deformable convolutional layer used to extract the deep layer, $F_l(\cdot)$ represents the 1$\times$1 variability convolutional layer used to extract the shallow layer. Therefore, the embedded $X_{FRM}$ is injected into the downstream model to provide prior knowledge to recognize the input image, which can be expressed by the following equation:
{\footnotesize \begin{equation}
		\boldsymbol{X}(p)={MLP}\left({FRM}\left({FRM}\left(\boldsymbol{X}_0^{1}, \boldsymbol{X}_l^{1}\right),{FRM}\left(\boldsymbol{X}_0^{2}, \boldsymbol{X}_l^{2}\right)\right)\right).
\end{equation}}

The $ l \in\{1, \ldots, L\}$, and the original and deep information of the single modality are first fused by feature reuse, and then the $FRM$ results of the two modalities are fused in the same way for the classification of downstream models.


\section{Experiments and Analysis}
\label{Experiments}

\begin{figure*}[htbp]
	\centering
	\includegraphics[scale=0.265]{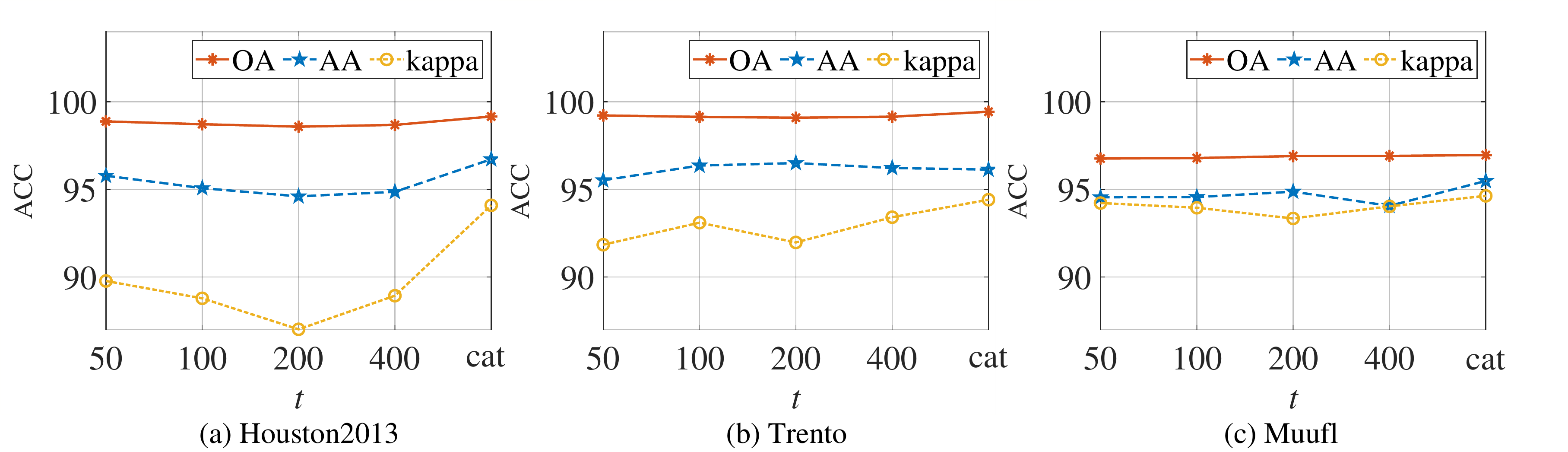}
	\centering
	\caption{The graph comparison between single step training and multi-step fusion training under different synchronization lengths is provided.}
	\label{zhexian}
\end{figure*}

\subsection{Experimental Settings}
\subsubsection{Datasets}To validate the effectiveness of the proposed method in analyzing high-dimensional data, three remote sensing multimodal datasets with hyperspectra, namely the Houston2013 dataset, the Trento dataset, and the Muufl dataset, are selected for verification of the proposed classification model. Detailed information about these datasets can be found in Table \ref{datadescripition}.

To assess the performance of the proposed method in classifying test images, three metrics are used: overall accuracy (OA), average accuracy (AA), and kappa ($\kappa$)  coefficient. OA measures the ratio of correctly classified test samples to the total number of test samples. AA represents the average accuracy across all classes. The $\kappa$ coefficient measures the agreement between the classification maps generated by the model and the true values provided.

 \begin{table}[htpb]
	\scriptsize 
	\setlength{\tabcolsep}{0.3mm}
	\renewcommand{\arraystretch}{1.1}
	\centering
	\begin{tabular}{c|c|c|c|c|c|c}
		\toprule[1.2pt]
		\textbf{Dataset}                  & \multicolumn{2}{c|}{\textbf{Houston2013}}           & \multicolumn{2}{c|}{\textbf{Trento}}          & \multicolumn{2}{c}{\textbf{Muufl}}                                                              \\
		\midrule[1.2pt]
		\textbf{Location}                 & \multicolumn{2}{c|}{Houston,Texas,USA} & \multicolumn{2}{c|}{Trento,Italy}    & \multicolumn{2}{c}{Long Beach,Mississippi,USA}                            \\
		\textbf{Sensor Type}     & ~~~~~HSI~~~~~      & DSM   & ~~HSI~~  & ~DSM~  & ~~~~~~~~HSI~~~~~~~~              &  DSM, DEM\\
		\textbf{Image Size (H)}            & 349               & 349               & 600             & 600            & 325                  & 325                     \\
		\textbf{Image Size (W)}           & 1905              & 1905              & 166             & 166            & 220                  & 220                     \\
		\textbf{Channel} & 144               & 1                 & 63               & 1                & 64               & 2                                   \\
		\bottomrule[1.2pt]   
	\end{tabular}
	\caption{Dataset Description.}
	\label{datadescripition}
\end{table}

\begin{figure*}[htbp]
	\centering
	\includegraphics[width=7in]{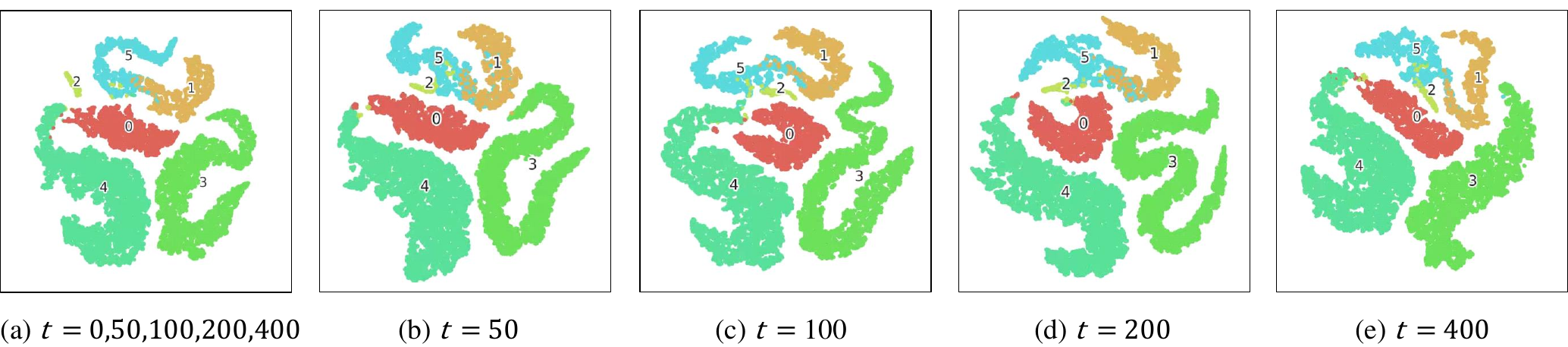}
	\caption{$t$-SNE effect display for different time steps.}
	\label{tsne}
\end{figure*}

\subsubsection{Implement details}
These experiments are conducted on a machine equipped with an NVIDIA A$100$ Tensor Core GPU. The training samples are randomly cropped to a size of $7 \times 7$. The Adam optimizer is employed with an initial learning rate set to 1e-3, and a weight decay of 5e-3 was applied. The training process spans $1000$ epochs. In addition, a step scheduler with a step size of $50$ and gamma value of $0.9$ is utilized. The batch size is set to $64$. Additionally, for incorporating noise, a total step size of $500$ was used, and the values of $t = 0, 50, 100, 200, 400$ are selected.

\begin{table*}[htbp]
	\centering
	\renewcommand{\arraystretch}{1.3}
	\setlength{\tabcolsep}{2.6mm}{
		\begin{tabular}{c|ccc|ccc|ccc}
			\toprule[1.2pt]
			\textbf{} & \textbf{} & \textbf{houston2013} & \textbf{} & \textbf{} & \textbf{Trento} & \textbf{} & \textbf{} & \textbf{MUUFL} & \textbf{ } \\ 
			\midrule[1.2pt] 
			\textbf{Method} & \textbf{OA($\%$)} & \textbf{AA($\%$)} & \textbf{$\kappa(\times100) $} & \textbf{OA($\%$)} & \textbf{AA($\%$)} & \textbf{$\kappa(\times100) $} & \textbf{OA($\%$)} & \textbf{AA($\%$)} & \textbf{$\kappa(\times100) $}  \\
			\midrule[1.2pt] 
			\textbf{RNN} & 62.61 & 62.24 & 59.64 & 96.43 & 92.38 & 95.21 & 88.79 & 75.84 & 85.18\\ 
			\textbf{Cross} & 91.84 & 92.70 & 91.16 & 97.82 & 97.35 & 97.09 & 87.29 & 63.81 & 82.75\\ 
			\textbf{CNN-2D} & 92.30 & 92.69 & 91.65 & 98.65 & 94.58 & 98.19 & 91.88 & 78.44 & 89.22\\ 
			\textbf{CALC} & 88.97 & 90.78 & 88.06 & 94.62 & 91.33 & 92.81 & 93.94 & 74.09 & 92.00\\ 
			\textbf{ViT} & 85.05 & 86.83 & 83.84 & 96.47 & 94.56 & 95.28 & 92.15 & 78.50 & 89.56\\ 
			\textbf{MFT} & 89.80 & 91.54 & 88.93 & 98.32 & 95.98 & 97.75 & 94.34 & 81.48 & 92.51\\ 
			\textbf{MDFL} & \textbf{99.16} & \textbf{96.72} & \textbf{94.09} & \textbf{99.43} & \textbf{96.13} & \textbf{94.41} & \textbf{96.96} & \textbf{82.63} & \textbf{94.63}\\ 
			\bottomrule[1.2pt]
	\end{tabular}}
	\caption{OA, AA and $\kappa$ on the Houston2013 dataset, Trento dataset and Muffl dataset by considering HSI and LiDAR data.}
	\label{com_result}
\end{table*}
 
\subsection{Ablation Study}
In this section, we conduct an ablation study to assess the individual contributions of different components in the proposed MDFL to MML. Specifically, five scenarios are designed: (A) a comparison of noise with various synchronization lengths selected by us against single-step training and multi-step fusion training; (B) the omission of frequency domain analysis during the training of diffusion models; (C) the exclusion of the original information in upstream model training from participating in feature reuse; (D) the consideration of the complete proposal model, MDFL.

\begin{table}[htbp]
	\centering
	\renewcommand{\arraystretch}{1.3}
	\setlength{\tabcolsep}{1.63mm}{
		\begin{tabular}{c|c|c|c}
			\toprule[1.2pt]
			\textbf{Description } & \textbf{OA($\%$)} & \textbf{AA($\%$)} & \textbf{$\kappa$($\times100$)} \\ 
			\midrule[1.2pt] 
			\textbf{(B)} & 97.75 & 95.62 & 93.24  \\ 
			\textbf{(C)} & 98.10 & 90.44 & 89.39  \\ 
			\textbf{(D)} & \textbf{98.51} & \textbf{91.82} & \textbf{94.37}  \\ 
			\bottomrule[1.2pt]
	\end{tabular}}
	\caption{Average results of ablation study for MDFL conduct on $3$ datasets. The best result is highlighted.}
	\label{Ablation}
\end{table}

As shown in Figure \ref{zhexian}, we observe with the line plots of three datasets that single-step training performs worse than multi-step fusion training in the process of forward diffusion, with validates the effectiveness of multi-step noise fusion. To further validate the benefits of multi-step time step fusion to classification, we demonstrate this with t-SNE maps at different time steps, as shown in Figure \ref{tsne}. It can be observed that (a) exhibits excellent clustering performance, with class 2 clearly separated from other classes in the two-dimensional projection space. However, in (b), (c), (d), and (e), class 2 shows incorrect associations with other classes in the space, with confirms the effectiveness of multi-step fusion. Additionally, Table \ref{Ablation} demonstrates that the average OA precision of MDFL across the three datasets is 98.51$\%$. In (B), by removing the joint analysis in the frequency domain, we observe a decrease in the results to 97.75$\%$, resulting in a performance loss of 0.76$\%$, which indicates the importance of multi-domain joint learning. The effectiveness of feature reuse in high-dimensional manifold structures is validated in (C). Therefore, in high-dimensional feature extraction, multi-objective domain joint learning and feature reuse modules can effectively enhance its performance.

\subsection{Comparisons with Previous Methods}

The accuracy performance of the proposed model as well as other models on the Houston2013, Trento, and Muufl datasets is presented in Table \ref{com_result}. The best results are denoted in bold. In selecting comparison methods, we consider classic deep learning techniques such as RNN \cite{cho2014properties}, mainstream RS multi-modal methods including Cross \cite{hong2020more}, CNN-2D and CALC \cite{9926173}, and transformer methods such as ViT \cite{dosovitskiy2021an} and MFT \cite{roy2023multimodal}. Our evaluation demonstrates that the proposed method outperforms other methods by achieving the highest OA, AA, and $\kappa$ across most classification tasks.

Our evaluation demonstrates that the proposed method attains the highest OA, AA, and $\kappa$ scores on most classification tasks, thus surpassing other methods. Specifically, on the Houston2013 dataset, MDFL exhibits superior accuracy across all categories, outperforming mainstream methods in terms of these coefficients. Although traditional methods perform better in terms of accuracy when combining HSI and LiDAR data, our model still presents significant advancements compared to other models. In comparison to transformer architecture, MDFL demonstrates a 9.36$\%$ increase in OA, a 5.18$\%$ increase in AA, and a 5.16$\%$ increase in $\kappa$ with MFT. When compared to Cross, MFT increases by 7.32$\%$ in OA, 4.02$\%$ in AA, and 2.93$\%$ in $\kappa$. Overall, MDFL achieves comprehensive performance improvements across three multimodal datasets and offers an effective solution for feature extraction from high-dimensional data.

\subsection{Result Visualization}

\begin{figure}[htbp]
	\centering
	\includegraphics[scale=0.635]{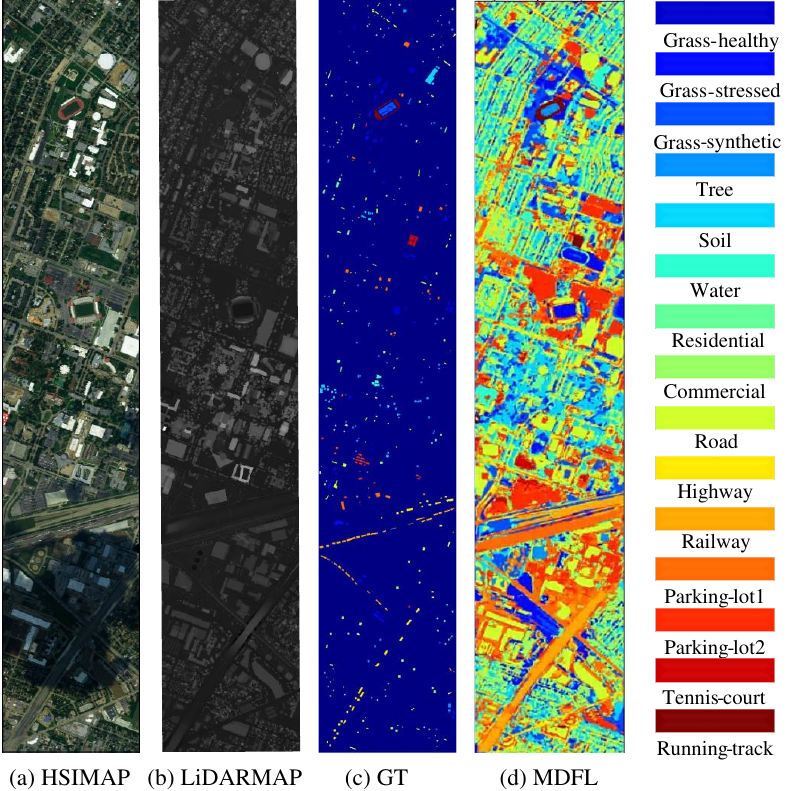}
	\centering
	\caption{Visualization of false-color HSI and LiDAR images using MDFL based on the Houston2013 dataset.}
	\label{vish}
\end{figure}

\begin{figure}[htbp]
	\centering
	\includegraphics[scale=0.748]{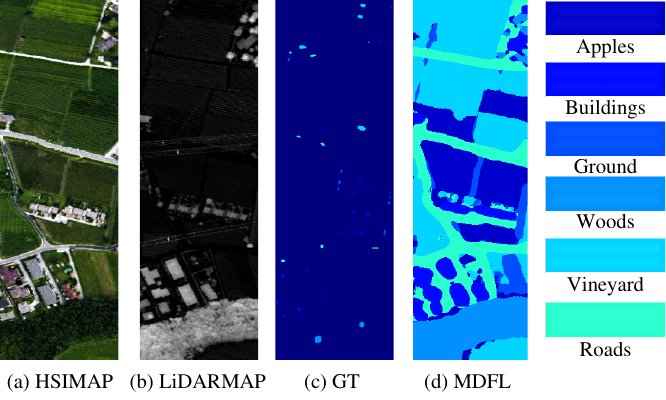}
	\centering
	\caption{Visualization of false-color HSI and LiDAR images using MDFL based on the Trento dataset.}
	\label{vism}
\end{figure}

\begin{figure}[htbp]
	\centering
	\includegraphics[scale=0.55]{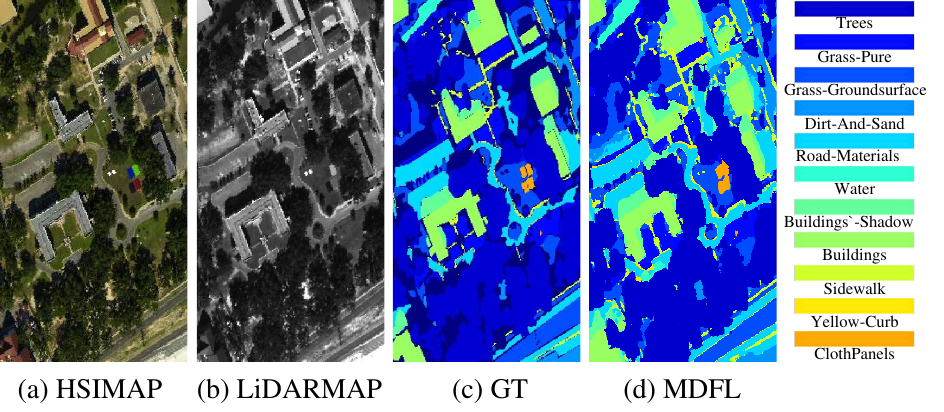}
	\centering
	\caption{Visualization of false-color HSI and LiDAR images using MDFL based on the Muffl dataset.}
	\label{vist}
\end{figure}

The results are depicted in Figure \ref{vish}, Figure \ref{vism}, and Figure \ref{vist}, illustrating the macroscopic performance evaluation of the classification graphs produced by MDFL. To achieve this, we employ a visualization technique that assigns a unique color to each class. MDFL reconstructs high-dimensional features by integrating the joint information from various domains, including spectral, spatial, and frequency domains. This approach effectively reduces the granularity of textures in the classification graphs, resulting in a more diverse and finely-detailed representation. Overall, MDFL is well-suited for generating classification maps with enhanced performance and intricate details, thus, it is particularly suitable for land use and scene classification applications.

\section{Conclusion}
In the context of high-dimensional image feature extraction, a significant challenge lies in developing a unified MMLmodel capable of facilitating joint information exchange and feature extraction across multiple domains. This problem has not been fully addressed within the previous framework that primarily focuses on a single spatial domain. Taking inspiration from the masking texture effect observed in the human visual system, we propose a novel approach for joint feature learning in high-dimensional multimodal images. This approach encompasses spectral domain, spatial domain, and frequency domain to comprehensively capture the intricate features. Additionally, we consider the manifold structure characteristics of high-dimensional data and introduce a feature reuse mechanism to aggregate deep and primitive features from multiple modes. With multi-domain joint learning, our method not only reveals the underlying patterns and structures of high-dimensional data, but also maximizes the performance of feature extraction. Extensive experiments have been conducted to validate the efficacy of our proposed method, MDFL, in tackling MML problems. The results consistently demonstrate that MDFL outperforms existing methods, showcasing superior performance and highlighting its potential in addressing the challenges of high-dimensional data analysis.
\label{Conclusion}

\bibliography{aaai23}

\begin{thebibliography}{26}
\providecommand{\natexlab}[1]{#1}

\bibitem[{Baranchuk et~al.(2021)Baranchuk, Rubachev, Voynov, Khrulkov, and
  Babenko}]{baranchuk2021label}
Baranchuk, D.; Rubachev, I.; Voynov, A.; Khrulkov, V.; and Babenko, A. 2021.
\newblock Label-efficient semantic segmentation with diffusion models.
\newblock \emph{arXiv preprint arXiv:2112.03126}.

\bibitem[{Cho et~al.(2014)Cho, Van~Merri{\"e}nboer, Bahdanau, and
  Bengio}]{cho2014properties}
Cho, K.; Van~Merri{\"e}nboer, B.; Bahdanau, D.; and Bengio, Y. 2014.
\newblock On the properties of neural machine translation: Encoder-decoder
  approaches.
\newblock \emph{arXiv preprint arXiv:1409.1259}.

\bibitem[{Croitoru et~al.(2023)Croitoru, Hondru, Ionescu, and
  Shah}]{croitoru2023diffusion}
Croitoru, F.-A.; Hondru, V.; Ionescu, R.~T.; and Shah, M. 2023.
\newblock Diffusion models in vision: A survey.
\newblock \emph{IEEE Transactions on Pattern Analysis and Machine Intelligence
  (TPAMI)}.

\bibitem[{Ding et~al.(2022)Ding, Lu, Fu, Li, and Ma}]{9926173}
Ding, K.; Lu, T.; Fu, W.; Li, S.; and Ma, F. 2022.
\newblock Global–Local Transformer Network for HSI and LiDAR Data Joint
  Classification.
\newblock \emph{IEEE Transactions on Geoscience and Remote Sensing (TGRS)}, 60:
  1--13.

\bibitem[{Dosovitskiy et~al.(2021)Dosovitskiy, Beyer, Kolesnikov, Weissenborn,
  Zhai, Unterthiner, Dehghani, Minderer, Heigold, Gelly, Uszkoreit, and
  Houlsby}]{dosovitskiy2021an}
Dosovitskiy, A.; Beyer, L.; Kolesnikov, A.; Weissenborn, D.; Zhai, X.;
  Unterthiner, T.; Dehghani, M.; Minderer, M.; Heigold, G.; Gelly, S.;
  Uszkoreit, J.; and Houlsby, N. 2021.
\newblock An Image is Worth 16x16 Words: Transformers for Image Recognition at
  Scale.
\newblock In \emph{International Conference on Learning Representations
  (ICLR)}.

\bibitem[{Esser, Rombach, and Ommer(2021)}]{esser2021taming}
Esser, P.; Rombach, R.; and Ommer, B. 2021.
\newblock Taming transformers for high-resolution image synthesis.
\newblock In \emph{Proceedings of the IEEE/CVF Conference on Computer Vision
  and Pattern Recognition (CVPR)}, 12873--12883.

\bibitem[{Fan et~al.(2021)Fan, Xiong, Mangalam, Li, Yan, Malik, and
  Feichtenhofer}]{fan2021multiscale}
Fan, H.; Xiong, B.; Mangalam, K.; Li, Y.; Yan, Z.; Malik, J.; and
  Feichtenhofer, C. 2021.
\newblock Multiscale vision transformers.
\newblock In \emph{Proceedings of the IEEE/CVF international conference on
  computer vision}, 6824--6835.

\bibitem[{Feng et~al.(2023)Feng, Song, Yang, Zhang, and Jiao}]{feng2023cross}
Feng, Z.; Song, L.; Yang, S.; Zhang, X.; and Jiao, L. 2023.
\newblock Cross-Modal Contrastive Learning for Remote Sensing Image
  Classification.
\newblock \emph{IEEE Transactions on Geoscience and Remote Sensing (TGRS)}.

\bibitem[{Han, Zheng, and Zhou(2022)}]{han2022card}
Han, X.; Zheng, H.; and Zhou, M. 2022.
\newblock Card: Classification and regression diffusion models.
\newblock \emph{Advances in Neural Information Processing Systems (NIPS)}, 35:
  18100--18115.

\bibitem[{Hong et~al.(2022)Hong, Gao, Hang, Zhang, and Chanussot}]{9179756}
Hong, D.; Gao, L.; Hang, R.; Zhang, B.; and Chanussot, J. 2022.
\newblock Deep Encoder–Decoder Networks for Classification of Hyperspectral
  and LiDAR Data.
\newblock \emph{IEEE Geoscience and Remote Sensing Letters (GRSL)}, 19: 1--5.

\bibitem[{Hong et~al.(2020)Hong, Gao, Yokoya, Yao, Chanussot, Du, and
  Zhang}]{hong2020more}
Hong, D.; Gao, L.; Yokoya, N.; Yao, J.; Chanussot, J.; Du, Q.; and Zhang, B.
  2020.
\newblock More diverse means better: Multimodal deep learning meets
  remote-sensing imagery classification.
\newblock \emph{IEEE Transactions on Geoscience and Remote Sensing (TGRS)},
  59(5): 4340--4354.

\bibitem[{Hu et~al.(2022)Hu, Cai, Lin, Wang, Yuan, Zhang, Timofte, and
  Van~Gool}]{hu2022hdnet}
Hu, X.; Cai, Y.; Lin, J.; Wang, H.; Yuan, X.; Zhang, Y.; Timofte, R.; and
  Van~Gool, L. 2022.
\newblock Hdnet: High-resolution dual-domain learning for spectral compressive
  imaging.
\newblock In \emph{Proceedings of the IEEE/CVF Conference on Computer Vision
  and Pattern Recognition (CVPR)}, 17542--17551.

\bibitem[{Jiang et~al.(2021)Jiang, Xie, Lei, Jiang, and Li}]{jiang2021lren}
Jiang, K.; Xie, W.; Lei, J.; Jiang, T.; and Li, Y. 2021.
\newblock LREN: Low-rank embedded network for sample-free hyperspectral anomaly
  detection.
\newblock In \emph{Proceedings of the AAAI Conference on Artificial
  Intelligence (AAAI)}, volume~35, 4139--4146.

\bibitem[{Karras et~al.(2022)Karras, Aittala, Aila, and
  Laine}]{karras2022elucidating}
Karras, T.; Aittala, M.; Aila, T.; and Laine, S. 2022.
\newblock Elucidating the design space of diffusion-based generative models.
\newblock \emph{Advances in Neural Information Processing Systems (NIPS)}, 35:
  26565--26577.

\bibitem[{Li et~al.(2022{\natexlab{a}})Li, Ling, Kim, Kreis, Fidler, and
  Torralba}]{li2022bigdatasetgan}
Li, D.; Ling, H.; Kim, S.~W.; Kreis, K.; Fidler, S.; and Torralba, A.
  2022{\natexlab{a}}.
\newblock BigDatasetGAN: Synthesizing ImageNet with pixel-wise annotations.
\newblock In \emph{Proceedings of the IEEE/CVF Conference on Computer Vision
  and Pattern Recognition (CVPR)}, 21330--21340.

\bibitem[{Li et~al.(2022{\natexlab{b}})Li, Lin, Zhou, Qi, Wang, and
  Jia}]{li2022mat}
Li, W.; Lin, Z.; Zhou, K.; Qi, L.; Wang, Y.; and Jia, J. 2022{\natexlab{b}}.
\newblock Mat: Mask-aware transformer for large hole image inpainting.
\newblock In \emph{Proceedings of the IEEE/CVF Conference on Computer Vision
  and Pattern Recognition (CVPR)}, 10758--10768.

\bibitem[{Liang et~al.(2022)Liang, Hu, Feng, and He}]{liang2022dine}
Liang, J.; Hu, D.; Feng, J.; and He, R. 2022.
\newblock Dine: Domain adaptation from single and multiple black-box
  predictors.
\newblock In \emph{Proceedings of the IEEE/CVF Conference on Computer Vision
  and Pattern Recognition (CVPR)}, 8003--8013.

\bibitem[{Liu et~al.(2021)Liu, Wang, Liu, Gandomi, Daneshmand, Muhammad, and
  De~Albuquerque}]{liu2021human}
Liu, S.; Wang, S.; Liu, X.; Gandomi, A.~H.; Daneshmand, M.; Muhammad, K.; and
  De~Albuquerque, V. H.~C. 2021.
\newblock Human memory update strategy: a multi-layer template update mechanism
  for remote visual monitoring.
\newblock \emph{IEEE Transactions on Multimedia (TM)}, 23: 2188--2198.

\bibitem[{Nichol and Dhariwal(2021)}]{nichol2021improved}
Nichol, A.~Q.; and Dhariwal, P. 2021.
\newblock Improved denoising diffusion probabilistic models.
\newblock In \emph{International Conference on Machine Learning (ICML)},
  8162--8171. PMLR.

\bibitem[{Pan et~al.(2019)Pan, Hartley, Liu, and Dai}]{pan2019phase}
Pan, L.; Hartley, R.; Liu, M.; and Dai, Y. 2019.
\newblock Phase-only image based kernel estimation for single image blind
  deblurring.
\newblock In \emph{Proceedings of the IEEE/CVF Conference on Computer Vision
  and Pattern Recognition (CVPR)}, 6034--6043.

\bibitem[{Roy et~al.(2023)Roy, Deria, Hong, Rasti, Plaza, and
  Chanussot}]{roy2023multimodal}
Roy, S.~K.; Deria, A.; Hong, D.; Rasti, B.; Plaza, A.; and Chanussot, J. 2023.
\newblock Multimodal fusion transformer for remote sensing image
  classification.
\newblock \emph{IEEE Transactions on Geoscience and Remote Sensing (TGRS)}.

\bibitem[{Song et~al.(2020)Song, Wei, Shu, Song, and Lu}]{song2020bi}
Song, K.; Wei, X.-S.; Shu, X.; Song, R.-J.; and Lu, J. 2020.
\newblock Bi-modal progressive mask attention for fine-grained recognition.
\newblock \emph{IEEE Transactions on Image Processing (TIP)}, 29: 7006--7018.

\bibitem[{Yang et~al.(2022)Yang, Zhang, Song, Hong, Xu, Zhao, Shao, Zhang, Cui,
  and Yang}]{yang2022diffusion}
Yang, L.; Zhang, Z.; Song, Y.; Hong, S.; Xu, R.; Zhao, Y.; Shao, Y.; Zhang, W.;
  Cui, B.; and Yang, M.-H. 2022.
\newblock Diffusion models: A comprehensive survey of methods and applications.
\newblock \emph{arXiv preprint arXiv:2209.00796}.

\bibitem[{Yao et~al.(2019)Yao, Pan, Li, and Mei}]{yao2019hierarchy}
Yao, T.; Pan, Y.; Li, Y.; and Mei, T. 2019.
\newblock Hierarchy parsing for image captioning.
\newblock In \emph{Proceedings of the IEEE/CVF International Conference on
  Computer Vision (ICCV)}, 2621--2629.

\bibitem[{Zhao et~al.(2023)Zhao, Bai, Zhu, Zhang, Xu, Zhang, Zhang, Meng,
  Timofte, and Van~Gool}]{zhao2023ddfm}
Zhao, Z.; Bai, H.; Zhu, Y.; Zhang, J.; Xu, S.; Zhang, Y.; Zhang, K.; Meng, D.;
  Timofte, R.; and Van~Gool, L. 2023.
\newblock DDFM: denoising diffusion model for multi-modality image fusion.
\newblock \emph{arXiv preprint arXiv:2303.06840}.

\bibitem[{Zhou et~al.(2023)Zhou, Sheng, Fan, Ye, He, Wang, and
  Chen}]{zhou2023hyperspectral}
Zhou, J.; Sheng, J.; Fan, J.; Ye, P.; He, T.; Wang, B.; and Chen, T. 2023.
\newblock When Hyperspectral Image Classification Meets Diffusion Models: An
  Unsupervised Feature Learning Framework.
\newblock \emph{arXiv preprint arXiv:2306.08964}.

\end{thebibliography}


\end{document}